\documentclass[letterpaper, 10 pt, conference]{ieeeconf}  
\IEEEoverridecommandlockouts                              

\usepackage{graphics} 
\usepackage{epsfig} 
\usepackage{times} 
\usepackage{amsmath} 
\usepackage{amssymb}  
\usepackage{tablefootnote}

\usepackage{cite}
\usepackage{amsmath,amssymb,amsfonts}
\usepackage{graphicx}
\usepackage{textcomp}
\usepackage{url}
\usepackage{tablefootnote}
\usepackage[table]{xcolor}
\usepackage{subcaption}
\usepackage{dsfont}

\newcommand{\modelname}{\text{TIP} }
\newcommand{\modelnamenospace}{\text{TIP}}

\title{\LARGE \bf
TIP: Task-Informed Motion Prediction for Intelligent Vehicles
}

\author{Xin Huang$^{1}$, Guy Rosman$^{2}$, Ashkan Jasour$^{1}$, Stephen G. McGill$^{2}$, John J. Leonard$^{1,2}$, Brian C. Williams$^{1}$
\thanks{$^{1}$Computer Science and Artificial Intelligence Laboratory, Massachusetts Institute of Technology, Cambridge, MA 01239, USA
        {\tt\footnotesize xhuang@csail.mit.edu }}%
\thanks{$^{2}$Toyota Research Institute, Cambridge, MA 02139, USA }%
}

\begin{document}

\maketitle
\thispagestyle{empty}
\pagestyle{empty}

\begin{abstract}
When predicting trajectories of road agents, motion predictors usually approximate the future distribution by a limited number of samples. This constraint requires the predictors to generate samples that best support the task given task specifications. However, existing predictors are often optimized and evaluated via task-agnostic measures without accounting for the use of predictions in downstream tasks, and thus could result in sub-optimal task performance.

In this paper, we propose a task-informed motion prediction model that better supports the tasks through its predictions, by jointly reasoning about prediction accuracy and the utility of the downstream tasks, which is commonly used to evaluate the task performance.
The task utility function does not require the full task information, but rather a specification of the utility of the task, resulting in predictors that serve a wide range of downstream tasks.
We demonstrate our approach on two use cases of common decision making tasks and their utility functions, in the context of autonomous driving and parallel autonomy.
Experiment results show that our predictor produces accurate predictions that improve the task performance by a large margin in both tasks when compared to task-agnostic baselines on the Waymo Open Motion dataset.

\end{abstract}

\section{INTRODUCTION}
Motion prediction is crucial for intelligent systems. It captures the distribution of future behavior of nearby road agents, and allows intelligent systems to plan and act.
The output distribution is often approximated via a set of \emph{weighted samples}~\cite{chai2019multipath,liang2020learning,huang2020diversitygan,gu2021densetnt}. The samples allow the downstream tasks to evaluate the predictions for risk assessment more efficiently, compared to complicated spatial distributions~\cite{wang2020fast}; the weights are necessary to provide an accurate estimate of the risk. In many cases, a predictor only affords a small set of trajectory samples, due to the time complexity of processing the predictions. For instance, evaluating one prediction sample for risk assessment may take up to a few milliseconds~\cite{wang2020fast}, making it impractical to evaluate a large number of samples for real-time decision making.

Traditionally, distance-based measures, such as displacement errors, are used to optimize and evaluate the prediction samples, and have proved tremendously useful for reducing and gauging prediction errors.
However, they do not account for the relevant downstream task, such as a planner that selects safe and efficient plans by reasoning about the predictions, and a driver assistance system that warns the driver when detecting risky behaviors.

\begin{figure}[t!]
    \centering
    \includegraphics[width=0.4\textwidth]{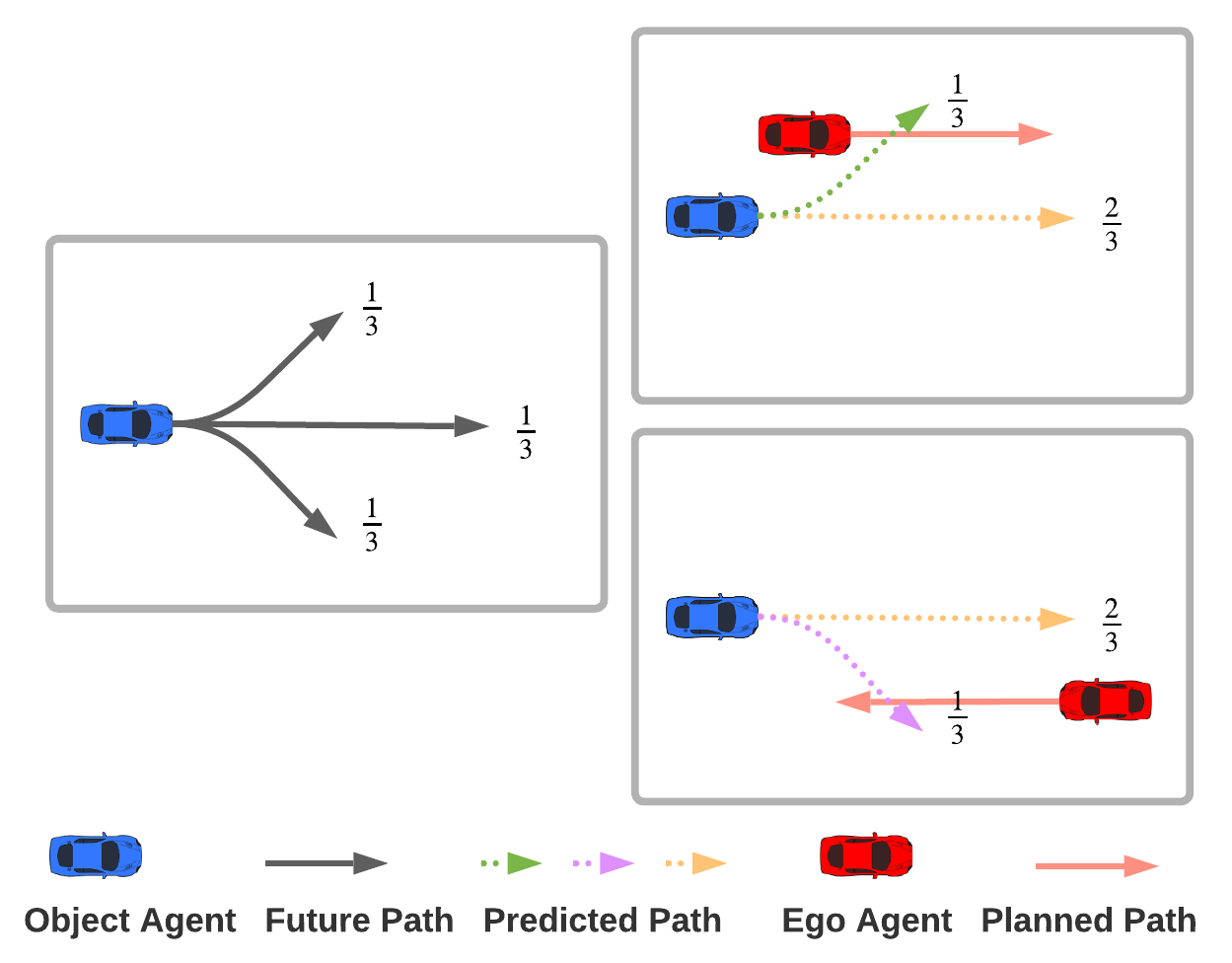}
    \caption{A motivating example for task-informed prediction, where three equally likely future trajectories for the object agent are illustrated on the left. Given a sample limit of two, a task-informed predictor supports the task by outputting different sets of prediction samples weighted by their importance to the task, depending on the task information, such as the future plan of the ego agent, as illustrated in the two scenarios on the right. In contrast, a task-agnostic predictor tells no difference between these two sample sets and may generate predictions leading to unsafe events.}
    \label{fig:motivating}
\vspace{-4mm}
\end{figure}

Consider a motivating example in Fig.~\ref{fig:motivating}, where the \textit{object agent}, defined as a nearby road agent of interest, may follow three possible future trajectories with an equal chance, as shown on the left. Due to the sample constraint, a predictor in this simplified example is only allowed to generate two trajectory samples\footnote{In practice, a predictor affords more than two samples, but from a much larger prediction space.}. We present two possible prediction outcomes on the right, denoted by the colored dashed lines. There are no differences between these two outcomes when evaluated by a task-agnostic distance-based metric, such as displacement error. However, when the predictions are used in a planning system for the \textit{ego agent} that aims to navigate safely with the object agent, one outcome may be favored over the other, depending on the ego plan. In the top right scenario, the green and the orange samples are more informative as they help the ego agent identify a potential near collision, whereas generating the purple and the orange predictions in this scenario may lead to an unsafe ego plan.
The purple and the orange samples are, however, favored in the bottom right scenario, given a different ego plan.
Therefore, it is important to reason about the downstream task when providing a limited set of prediction samples. 
The predictor should also provide an accurate estimate of the importance (or weight) of each sample with respect to the task. As we show in both scenarios, the predictor estimates the weight of each sample to help the planner accurately compute the probability of collision with the object agent.

In short, a predictor should provide a faithful estimate of the future trajectories of nearby road agents through a limited number of weighted samples.
These predictions should assist decision making by characterizing nearby road agents and their future actions~\cite{worrall2012context}, providing an \emph{approximate sufficient statistics}~\cite{chen2020neural} for the task.
Furthermore, its design should accommodate a variety of decision making tasks as opposed to a specific task, for flexibility in broader applications.

With this rationale in mind, we propose \emph{\modelnamenospace}, a \emph{Task-Informed motion Predictor}, which is learned by jointly optimizing prediction accuracy and the performance of downstream tasks. 
The training loss leverages a specification of the task, such as its utility function, instead of ignoring the task or requiring the task itself to be co-trained (e.g. optimizing a specific planner with the predictor).
Compared to existing prediction methods, \modelname generates predictions that improve the downstream task performance by providing relevant information to the task given limited samples.
In addition, it accommodates a variety of decision making tasks within intelligent systems, through different utility functions that characterize the tasks, compared to models that are constrained to a specific task and require that task to be differentiable~\cite{zeng2019end}. 
While TIP is designed to support an arbitrary decision making task given their utility functions, we present two common decision making tasks, in the context of autonomous driving and parallel autonomy, to demonstrate the effectiveness of our proposed model.

Our contributions are as follows: 
i) We present a task-informed motion prediction model for intelligent systems that can be used to improve the performance of downstream tasks. Our approach covers a wide range of tasks, through utility function surrogates that characterize the tasks. 
ii) We show two case studies of our predictor being used in a planning task and a warning task, which are commonly used by state-of-the-art intelligent systems such as autonomous driving and parallel autonomy.
iii) We demonstrate that our system helps achieve better task performance across both tasks through detailed quantitative and qualitative comparisons to task-agnostic baselines in a large-scale motion prediction dataset. 

\section{RELATED WORK}

\subsection{Motion Prediction}
Motion prediction has been studied extensively in the last few years, from predicting vehicles~\cite{wiest2012probabilistic,kim2017probabilistic,huang2021carpal} to vulnerable road users such as pedestrians and cyclists~\cite{alahi2016social,gu2021densetnt,sun2022m2i}. 
Due to the complexity in multi-modal future trajectory distributions, motion predictors often approximate the output distribution through simplified representations, such as a weighted set of samples~\cite{huang2020diversitygan,gu2021densetnt}, Gaussian mixture models~\cite{wiest2012probabilistic,chai2019multipath}, and occupancy maps~\cite{kim2017probabilistic}. We refer to~\cite{mozaffari2020deep} for a more comprehensive list of literature. In this work, we focus on sampling-based predictors, which offer a good balance between retaining enough continuous information and allowing the downstream tasks to efficiently process the predictions as deterministic positions.

Despite the recent efforts on supplying additional labels to improve prediction accuracy~\cite{deo2018convolutional,chai2019multipath,mercat2020multi}, most existing literature quantifies prediction performance through task-agnostic measures, including distance-based metrics such as minimum average displacement error (ADE)~\cite{alahi2016social} and distribution-based metrics such as negative log-likelihood (NLL)~\cite{wiest2012probabilistic}. However, such metrics do not account for the use of predictions in the downstream task. In other words, predictions with the same accuracy may lead to different outcomes for tasks such as planning, as illustrated in Fig.~\ref{fig:motivating}. Therefore, we propose a prediction model that accounts for not only prediction accuracy, but also the utility of the downstream task given the predictions, to allow better integration between predictions and the task.

A concurrent work~\cite{ivanovic2021injecting} is proposing task-aware metrics to evaluate the performance of motion predictors. It presents a proof-of-concept metric that favors the prediction samples based on the sensitivity to the ego agent's plan. Results show that the proposed metric better provides a measure of planner performance compared to distance-based metrics.
Different from ~\cite{ivanovic2021injecting} that focuses on proposing task-aware metrics to \emph{evaluate} predictor performance, our method goes beyond by leveraging task-specific information to \emph{train} a motion predictor. 
This allows us to maximize the value of predictions for downstream tasks, as we demonstrate in realistic driving scenarios. More importantly, by training with a task-informed loss, our predictions cover approximate sufficient statistics~\cite{jiang2017learning,chen2020neural} of the nearby road agents, and leverage even imperfect information of downstream tasks.

\subsection{Prediction for Tasks}
Existing works leverage a learned motion predictor to support a variety of tasks, including risk assessment~\cite{wang2020fast}, driver safety detection~\cite{huang2021carpal}, and most commonly, planning for autonomous systems~\cite{schmerling2018multimodal,sadat2019jointly,schaefer2020leveraging,nishimura2020risk,eiffert2020path,casas2021mp3}.
They often decouple the optimization of the predictor and the optimization of the task. As a result, the predictor is unaware of its influence on the downstream task and its predictions may not be informative for the downstream task. In this work, we present a more effective predictor that is optimized directly through the utility of the downstream task.

A prior work~\cite{huang2021carpal} proposes a multi-task predictor that approximates the utility and its uncertainties of the downstream driver safety detection task, yet the predictor is still learned with a single objective of optimizing the accuracy. In contrast, our model integrates the task utility into the trajectory prediction directly, allowing the prediction results to better support the task.

Our work is closely related to \emph{prediction and planning} (P\&P) literature~\cite{zeng2019end}, which jointly optimizes the prediction module and the planning module. Compared to end-to-end planning models~\cite{bojarski2016end,bansal2018chauffeurnet,codevilla2018end} that generate planning results from raw sensor inputs, P\&P, as a more structured approach, provides better interpretability in its decision making process, and achieves better planning performance~\cite{zeng2019end}.
One limitation of P\&P is that it requires a fully differentiable pipeline that includes a differentiable planner. 
In many practical planning systems that involve black-box modules or off-the-shelf components ~\cite{bacha2008odin,kuwata2008motion}, it is a nontrivial task to differentiate the planner.
In contrast, our prediction model only requires the utility function that characterizes the task.
The utility function is planner-agnostic, allowing the predictor to support a family of planning algorithms, instead of a specific planner as in P\&P. Furthermore, a differentiable utility function is usually easier to acquire than a differentiable planner. This makes our model more broadly applicable to non-differentiable planners and tasks that are more than planners, as we show in our experiments.

Instead of assuming a specific differentiable planner,~\cite{tang2019multiple} generates joint predictions for both the ego agent and the object agents, and uses the ego prediction for planning. Such approach offers a great advantage by simply imitating the future behavior of the ego agent without requiring a specific planner, yet the predicted trajectory for the ego agent is prune to input noise and modeling error, making it less reliable to be used for planning in practice.
\vspace{-3mm}
\section{PROBLEM FORMULATION}
\vspace{-2mm}
\textbf{Motion Prediction}
The prediction system takes input as i) task-specific input information $V$ and ii) observed agent trajectories $O = \{\mathbf{o}_t\}_{t=-T_p+1}^0$ over a fixed past horizon $T_p$, where $\mathbf{o}_t = [o_{1,t}, \ldots, o_{N,t}]$ includes continuous positions at time step $t$ for up to $N$ agents. The positions are normalized with respect to the center of the last observed positions of all agents.
The task input $V$ depends on the specific information from the task, such as an ego planned trajectory, as customary in conditional prediction approaches~\cite{tolstaya2021identifying,ngiam2021scene,salzmann2020trajectron++,song2020pip}.
The output is a weighted set of $K$ joint trajectory samples $\mathcal{S} = \{(w^{(k)}, \mathbf{x}^{(k)})\}_{k=1}^K$ for all agents, where $\mathbf{x}^{(k)} = \{\mathbf{x}_t^{(k)}\}_{t=1}^{T_f}$ denotes future trajectory sequences of all agents, i.e. $\mathbf{x}_t^{(k)} = [x_{1,t}^{(k)}, \ldots, x_{N,t}^{(k)}]$, up to a fixed future horizon $T_f$.

\textbf{Utility-Based Decision Making}
The task-informed prediction aims to allow accurate estimates of task utility for an arbitrary downstream decision making task. The utility (or reward) serves as a quantitative measurement of task performance and is commonly used in modern decision making tasks.
In this work, we define the task specification as a tuple $(\mathcal{I}, u)$, where $\mathcal{I}$ is a set of candidate decisions for the task, such as plans of the ego agent; $u$ is a task-specific differentiable utility function mapping a decision $I \in \mathcal{I}$ and the task-informed predictions $\mathcal{S}$ to a scalar that quantitatively measures the performance of the decision. For the sake of simplicity, we define $u_I = u(I, \mathcal{S})$ in the rest of the paper. This specification allows us to optimize our model to support different tasks through their task-specific surrogate utility functions, without requiring a specific task pipeline. 


Finally, we define the task objective to obtain the optimal decision that maximizes the utility among all candidate decisions:
\begin{equation}
    I_{O} = \arg \max_{I \in \mathcal{I}} u_I.
\label{eq:task}
\end{equation}

At training time, we can obtain the utilities of all candidates and find the optimal decision using Eq.~\eqref{eq:task} given ground truth future trajectories. We optimize the model by maximizing the following $\textit{softmax}$ over the decision utilities:
\begin{equation}
    R_{task} = \frac{\exp(u_{I_{O}})}{\sum_{I' \in \mathcal{I}} \exp(u_{I'})},
\label{eq:task_reward}
\end{equation}
which is widely used in reinforcement learning~\cite{sutton2018reinforcement} to encourage the optimal decision to have higher utility compared to the other decisions.





\section{APPROACH}
\begin{figure}[t!]
\vspace{1mm}
    \centering
    \includegraphics[width=0.4\textwidth]{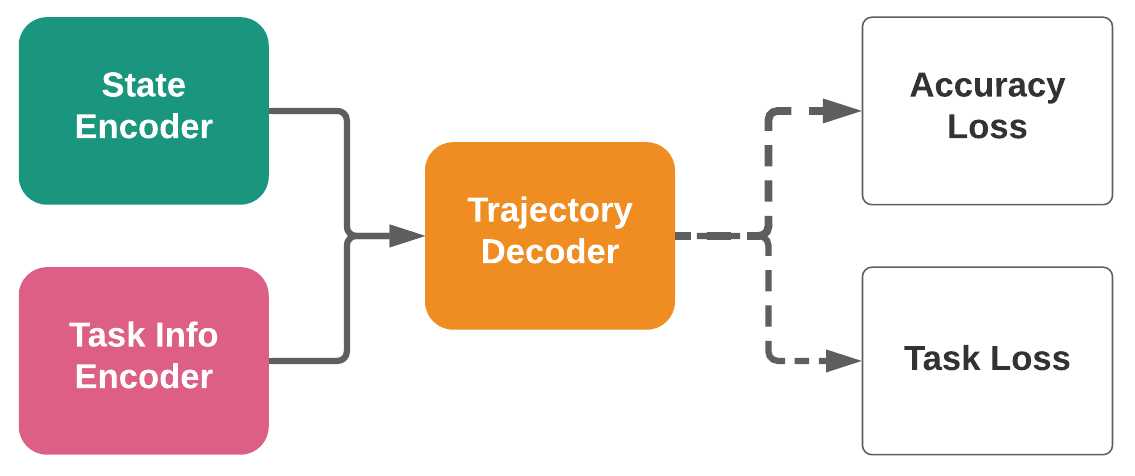}
    \caption{Diagram of the proposed task-informed prediction model, which includes a state encoder that encodes observed past agent states, a task information encoder that encodes additional task input, and a trajectory decoder that decodes future trajectory predictions. The model is trained through an accuracy loss term that optimizes prediction accuracy and a task loss term that guides the model to favor predictions supporting the downstream task.}
    \label{fig:model}
\end{figure}

Our predictor leverages an encoder-decoder model, as depicted in Fig.~\ref{fig:model}, following a standard architecture in~\cite{ettinger2021waymo}.
The state encoder encodes the observed trajectory for up to $N$ agents. For each agent $i$, an MLP is used to encode the position at each time step, and an LSTM is used to encode the position encodings from time step $-T_P + 1$ to 0 into a hidden state $h_{\mathcal{S}_i}$. The hidden states of each agent are concatenated into a joint hidden state $h_{\mathcal{S}}$. If there are fewer than $N$ agents in a scene or there exist invalid positions for any agent, we zero-mask the encodings of missing positions.
The task information encoder encodes task-specific inputs $V$, such as the future plan of the ego agent, through a separate model into a separate hidden state $h_V$. The structure of the task information encoder depends on the input representation, and we defer a detailed description in our experiments.
The trajectory decoder model takes the concatenated encoded states from both encoders, i.e. $h = h_{\mathcal{S}} \oplus h_V$, and predicts a set of $K$ joint trajectory samples $\mathcal{S}$ and their weights.

We train the model by jointly optimizing prediction accuracy and task performance,
\begin{equation}
    \mathcal{L} = \mathcal{L}_{acc} + \alpha \mathcal{L}_{task},
\label{eq:loss}
\end{equation}
where $\alpha$ determines the relative weight between two terms. In the experiments, we present the trade-off between prediction accuracy and task performance, by varying $\alpha$ values.

The accuracy loss $\mathcal{L}_{acc}$ measures the accuracy of the prediction samples compared to the ground truth future trajectory $\hat{S}$. We follow the standard motion prediction literature, i.e. \cite{cui2019multimodal,chai2019multipath}, based on a variant of variety loss proposed in~\cite{alahi2016social}:
\begin{equation}
    \mathcal{L}_{acc} = \sum_{k=1}^{K} \mathds{1}(k = \hat{k})(-\log w^{(k)} + ||\mathbf{x}^{(k)} - \hat{S}||_2),
\label{eq:loss_acc}
\end{equation}
where $\hat{k}$ is the index of the best prediction sample, in terms of $L2$ distance to the ground truth trajectory. 

The task term $\mathcal{L}_{task}$ minimizes the negation of Eq.~\eqref{eq:task_reward} so that the optimal decision would have higher utility compared to other candidate decisions given the predictions.
\begin{equation}
    \mathcal{L}_{task} = -R_{task}.
\label{eq:loss_task}
\end{equation}

While \modelname can be used for an arbitrary decision making task that requires behavior prediction given its task specifications, in the rest of this section, we present two use cases of tasks and their utility functions to demonstrate the flexibility of our approach over different tasks. We defer implementation details on both tasks in Section~\ref{sec:exp}.

\subsection{Example Task 1: Planning for Autonomous Systems}
In the planning task, we assume that the ego agent is equipped with an arbitrary planner that generates a set of $M$ motion plan candidates $\mathcal{I}_P = \{\tau_1, \ldots, \tau_M\}$.
The planning utility function is defined to capture safety and efficiency:
\begin{equation}
    u_P(\tau, \mathcal{S}) = u_{\text{efficiency}}(\tau) + \beta u_{\text{safety}}(\tau, \mathcal{S}),
\label{eq:r_p}
\end{equation}
where $\tau$ is an ego plan candidate, and $\mathcal{S} = \{(w^{(k)}, x_{\text{object}}^{(k)})\}_{k=1}^K$ is a weighted set of object prediction samples generated from our predictor. In this task, the predictor takes the input\footnote{We follow existing conditional behavior prediction literature \cite{salzmann2020trajectron++,tolstaya2021identifying} to generate predictions
conditioning on the future ego plan. Such predictors have demonstrated to predict reactive agent behaviors and improve accuracy, and are useful for a wide range of planners that generate initial candidate
plans using simplified prediction models, such as~\cite{frazzoli2002real,berg2011reciprocal}.} of the observed agent states and the ego plan $\tau$ as the task input $V$ to produce the prediction samples $\mathcal{S}$. The efficiency term $u_{\text{efficiency}}$ measures the travelled distance of the ego plan. The safety term $u_{\text{safety}}$ measures the expected closest distance between the ego plan and the object predictions, computed as follows:
\begin{equation}
    u_{\text{safety}}(\tau, \mathcal{S}) = \sum_{k=1}^{K} w^{(k)} \min_{t=1\ldots T_f} ||\tau_t - x_{\text{object},t}^{(k)}||_{2}.
\end{equation}

In practice, the improvement of the safety utility diminishes if the agents are far away from each other. Therefore, we upper bound the utility by a safety threshold $d_{\text{safe}}$:
\begin{equation}
    u_{\text{safety}}(\tau, \mathcal{S}) = \min(d_{\text{safe}}, \sum_{k=1}^{K} w^{(k)} \min_{t=1\ldots T_f} ||\tau_t - x_{\text{object},t}^{(k)}||_{2}).
\end{equation}
\vspace{-8mm}
\subsection{Example Task 2: Warning for Parallel Autonomy}
A pre-collision warning system is widely adopted in parallel autonomy (or shared autonomy), as a vehicle shared-control framework~\cite{saleh2013shared,schwarting2017parallel,huang2021carpal} that monitors driver actions and warns before an unsafe event could happen.
The warning system differs from the planning system in a few ways. First, it requires a \emph{joint} predictor for both ego agent and object agent, as the ego agent is controlled by a driver and the future path is unknown to the predictor. This requires predicting the joint behavior in the future to determine if a near collision is likely. 
Second, it provides \emph{no task-specific input} to the predictor, as it only sends a warning to the driver and does not induce any actual interactions with the world. As a result, the predictor produces prediction samples $\mathcal{S}$ by taking only the observed agent states as inputs.

The warning system is a binary decision making system that chooses an action from $\mathcal{I}_W = \{\texttt{warn}, \lnot\texttt{warn}\}$. 
The utility of a warning action is equivalent to the likelihood of near collision between the object agent and the ego agent. 
To compute the near collision likelihood, we follow the two-step procedure given the joint prediction samples $\mathcal{S} = \{(w^{(k)}, x_{\text{ego}}^{(k)}, x_{\text{object}}^{(k)})\}_{k=1}^K$ for the ago agent and the object agent.
First, the system computes the collision score $r^{(k)} \in \{0, 1\}$ as a Boolean value for each trajectory sample:
\begin{equation}
    r^{(k)} = \big(\min_{t=1\ldots T_f} ||x_{\text{ego},t}^{(k)} - x_{\text{object},t}^{(k)}||_2 < d_{\text{warn}}\big),
\label{eq:warn}
\end{equation}
where $d_{\text{warn}}$ is the minimum safety distance threshold allowed. The collision score is 1 if the closest distance between two agents is smaller than this threshold, and 0 otherwise.

Next, we compute the overall collision likelihood by taking the expected collision score $r$ as the weighted sum of individual warning scores: $u_W(\texttt{warn}) = r = \sum_{k=1}^K w^{(k)} r^{(k)}$.
Intuitively, the utility of $\lnot\texttt{warn}$ is the likelihood of no near collision, i.e.
$
    u_W(\lnot \texttt{warn}) = 1 - u_W(\texttt{warn}).
$

To compute the ground truth optimal decision, we compute the likelihood of near collision from the observed future trajectories following the same procedure in Eq.~\eqref{eq:warn}.
Since the observed future trajectories are deterministic, the resulting likelihood is either 0 or 1. 

\section{EXPERIMENTS}
In this section, we show experimental results in two different tasks to demonstrate the advantage of our proposed task-informed predictor on a naturalistic driving dataset. 

\label{sec:exp}
\subsection{Dataset}
We train and validate our model in the Waymo Open Motion dataset~\cite{ettinger2021waymo}. It is one of the largest motion prediction datasets in terms of the number of scenes, total time, and prediction horizon, i.e. 8 seconds with 80 time steps. 
More specifically, we focus on the interactive dataset that is mined to cover interesting interactions. This allows us to demonstrate the effectiveness of our model in complicated long-term interacting scenarios.
We follow the standard train/validation split from the dataset.

\subsection{Example Task 1: Planning for Autonomous Systems}
\label{sec:planner}

\subsubsection{Model Details}
\label{sec:plan_model}
The MLP in the state encoder has 32 neurons, followed by ReLU and dropout layers with a rate of 0.1. The LSTM has a hidden size of 32 and an output dimension of 32. 
The task information encoder encodes the planned trajectory of the ego agent, as the task-specific input $V$, through an MLP with 32 neurons, followed by ReLU and dropout layers with a rate of 0.1.
The trajectory decoder uses a two-layer MLP with 32 neurons to output $\mathcal{S}$ that includes the predicted trajectory samples and their weights. We choose $\alpha = 20$ to keep the two loss magnitudes on the same scale, and $\beta = 5$ to prioritize safe driving.
The model is trained for 20 epochs and is optimized using Adam~\cite{kingma2014adam}, with a batch size of 32 and a learning rate of $10^{-3}$. It takes approximately 2 milliseconds to generate all samples in each example.

\subsubsection{Task Details}
To simulate the planning task, we select the ego agent and the object agent randomly from the interactive pair in the Waymo data. The ego planner simulates three planned trajectories based on the observed future trajectory of the ego agent. It interpolates and scales the trajectory coordinates by 0.8x, 1.0x, and 1.2x at each time step to simulate conservative driving, normal driving, and aggressive driving, while imposing limits on acceleration and speed following~\cite{bokare2017acceleration}. This simple approach provides multiple driving options while ensuring the plans are realistic and closely follow the agent intention from data (see examples in Fig.~\ref{fig:planning_examples}), yet one can use an arbitrary planner to supply the plan candidates in practice.


In order to find the ground truth optimal plan, we also have to simulate the reactive behavior of the object agent in the future, which depends on the interaction type between two agents.
When the object agent is yielding to the ego agent in the data, we modify its future trajectory with an equal chance to either speed up to pass or slow down to keep yielding, in response to the \emph{conservative} ego plan. We keep its future trajectory unchanged in response to the other two ego plans.
In contrast, when the object agent is being yielded by the ego agent in the data, we modify its future trajectory with an equal chance to either slow down to yield or speed up to maintain the lead, in response to the \emph{aggressive} ego plan.
Simulating realistic object agent behavior for simulation purposes is a topic of ongoing research \cite{caesar2021nuplan}.
In our experiments, we follow a simple heuristic based on how humans normally react to others and find it to be realistic and effective. In addition, to validate the representativeness of our simulation model, we have experimented with an additional simulation model based on IDM, as in~\cite{bernhard2020bark}, and observed consistent improvements in task performance using our proposed approach. Full results are available in Table~\ref{tab:planning_idm}.

We choose the threshold $d_{\text{safe}}$ to be 3.64 meters as the 10\% percentile of the pairwise closest distances in the Waymo dataset. This value is smaller than the radius of a regular car with a length of approximately 4.5 meters and a width of approximately 2.0 meters~\cite{niroomand2021vehicle}.

\subsubsection{Quantitative Results}
We compare our proposed model \modelnamenospace$_P$, with a Task-Agnostic Predictor (we refer to it as TAP in the rest of the experiments) that uses the same model as ours, but is trained with only the \emph{accuracy loss} in Eq.~\eqref{eq:loss_acc}. 
This baseline is equivalent to the baseline model proposed in~\cite{ettinger2021waymo} and represents a broad prediction literature that ignores predictions in downstream tasks or decouples prediction and the tasks. 
In addition, we demonstrate that our proposed approach can be applied to a different utility function, \modelnamenospace$_{Pa}$, that is trained for the same planning task but represents a drastically different altruistic planner ($_{Pa}$):
\begin{equation}
    u_{Pa}(\tau, \mathcal{S}) = u_{\text{efficiency}}(\tau_{\text{object}}) + \beta u_{\text{safety}}(\tau, \mathcal{S}),
\label{eq:r_p}
\end{equation}
where $\tau_{\text{object}}$ is the simulated trajectory of the object agent that reacts to the ego plan $\tau$. This utility function models an altruistic planner that favors the object agent as opposed to the ego agent. Such a planner is commonly seen in robotics social navigation that minimally interferes with humans~\cite{schaefer2020leveraging}. 

We evaluate each model on prediction accuracy and task performance. The prediction accuracy is measured by minADE/minFDE metrics~\cite{alahi2016social}, as standard in motion prediction benchmarks~\cite{chang2019argoverse,ettinger2021waymo}. In addition, we present weighted ADE (wADE) and weighted FDE (wFDE) metrics that measure the expected errors given the predicted weights. The unit of all accuracy metrics is in meters.
We measure the task performance through recall and fall-out. 
Recall measures the percentage of optimal plans that are successfully recognized.
Fall-out measures the percentage of false alarms, i.e. sub-optimal plans that are wrongly recognized as optimal.
We plot the recall and fall-out at various thresholds, as receiver operating characteristic (ROC) curve~\cite{fawcett2006introduction}, and compute its area under the curve (AUC) score to determine the task performance. 
As our planner is dealing with a multi-class decision making problem, we compute the AUC-ROC score using the one-vs-one methodology, which is insensitive to data balance~\cite{fawcett2006introduction}.

\begin{table}[t!]
\vspace{2mm}
\centering
\begin{tabular}{c|cccc}
\hline
Model & minADE$\downarrow$ & minFDE$\downarrow$ & AUC-ROC$_P$$\uparrow$ & AUC-ROC$_{Pa}$$\uparrow$  \\ \hline
TAP     & \textbf{2.80}	& \textbf{6.44} & 0.594 & 0.616 \\
\modelnamenospace$_P$     & 2.89 &	6.54	&	\cellcolor{yellow!50}\textbf{0.667} & 0.586 \\
\modelnamenospace$_{Pa}$     & 2.93 &	6.51	&	0.555 & \cellcolor{yellow!50}\textbf{0.697} \\ \hline
\end{tabular}
\caption{Comparison between our TIP and the baseline model, in terms of prediction accuracy and task performance, on two tasks $P$ and $Pa$. The task-informed predictors trade off little accuracy to much better task performance. Our approach supports multiple utility functions, as suggested by the relevant metrics highlighted in the colored cells.}
\label{tab:planning}
\vspace{-4mm}
\end{table}

\begin{table}[t!]
\vspace{2mm}
\centering
\begin{tabular}{c|ccccc}
\hline
$\alpha$ & minADE$\downarrow$ & minFDE$\downarrow$ & wADE$\downarrow$ & wFDE$\downarrow$ & AUC-ROC$\uparrow$  \\ \hline
0     & \textbf{2.80} &	\textbf{6.44} & \textbf{5.76} & \textbf{14.32}    &		0.594 \\
1     & 2.83 &	6.47 & 5.82 & 14.46 &	0.613 \\
5     & 2.85 & 6.52 & 5.89 & 15.07 & 0.623 \\
20     & 2.89 &	6.54 & 6.66 & 16.69 &	0.667 \\
100     & 4.01 & 8.67 & 9.72 & 23.14 & \textbf{0.676} \\ \hline
\end{tabular}
\caption{Performance of TIP$_P$ as a function of the task loss coefficient $\alpha$.
}
\label{tab:planning_alpha}
\vspace{-4mm}
\end{table}

We report two separate AUC-ROC metrics, AUC-ROC$_P$ and AUC-ROC$_{Pa}$, depending on which utility function is used to determine the ground truth optimal plan. The results are summarized in Table~\ref{tab:planning}, where we color the cell to highlight the results measured by their relevant metrics. 

\textbf{Trade off accuracy for better task performance} 
While the task-agnostic model (TAP) achieves the best prediction accuracy, our model achieves much better task performance at the cost of little accuracy. More specifically, compared to TAP, \modelnamenospace$_P$ improves the task performance by 12.29\% at the cost of 1.55\% accuracy loss in terms of minFDE, and \modelnamenospace$_{Pa}$ improves the task performance by 13.15\% at the cost of 1.09\% accuracy loss.
In most cases, the task performance matters more than the prediction accuracy, as the planner interacts directly with the world and a small error may lead to undesirable outcomes. 
We present the trade-off between prediction accuracy and task performance by varying $\alpha$ values, as in Table~\ref{tab:planning_alpha}.

\textbf{Adapt to multiple utility functions} 
We show that our model can be adapted to multiple utility functions within the same planner task, through results highlighted by the yellow cell. For instance, when training with a different utility function $u_{Pa}$ that favors altruistic behavior, our model achieves good task performance on the relevant metric (e.g. \modelnamenospace$_{Pa}$ achieves a high AUC-ROC$_{Pa}$ score), although it tends to perform poorly on the drastically different task, as these two tasks adopt competing objectives on being ``selfish'' versus being ``altruistic''. In practice, given a specific task, the user can specify the utility function that best describes the task objective to serve their own need.

\subsubsection{Qualitative Results}
\label{sec:planning_qual}
\begin{figure}[t!]
\vspace{2mm}
    \centering
    \includegraphics[width=0.48\textwidth]{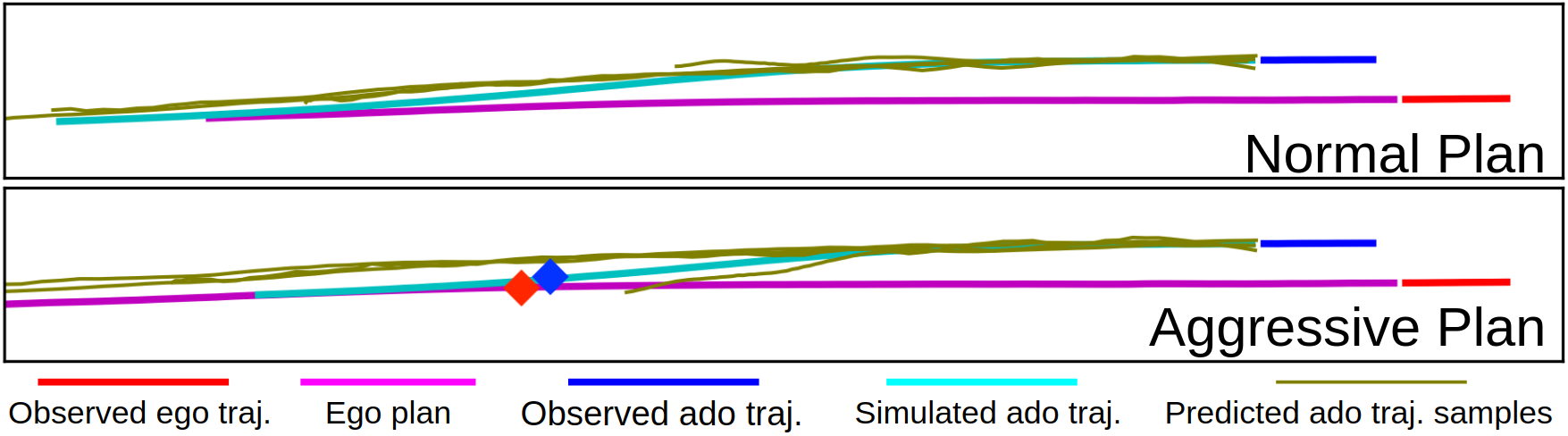}\\\vspace{-1mm}(a) Our proposed method \modelnamenospace$_P$\vspace{1mm}\\
    \includegraphics[width=0.48\textwidth]{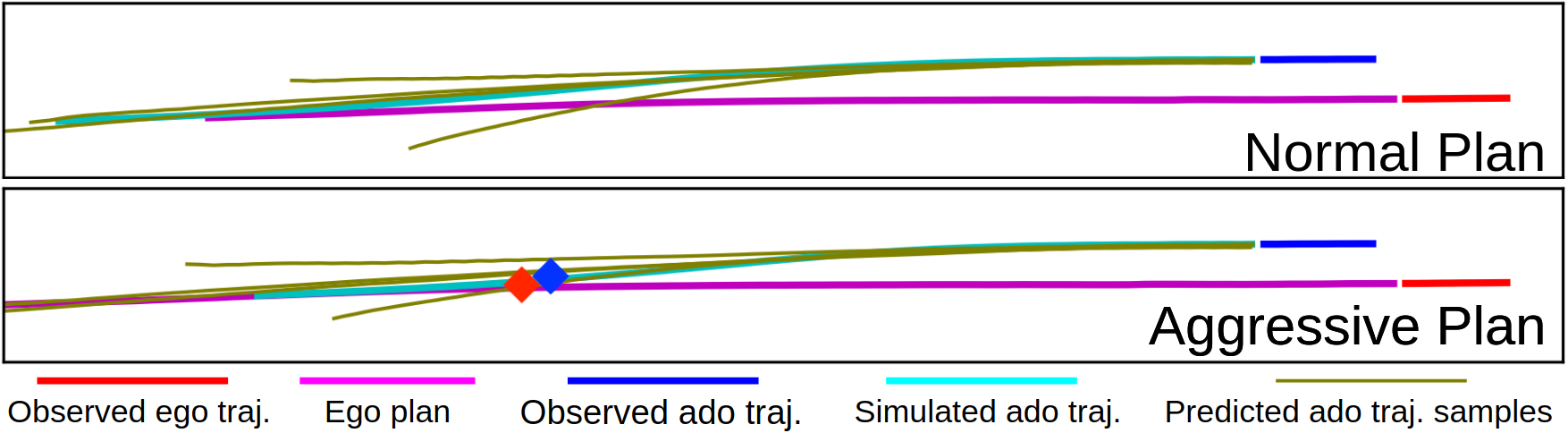}\\\vspace{-1mm}(b) Task-agnostic baseline TAP
    \caption{Predictions from \modelnamenospace$_P$ (a) and TAP (b) in a representative example, where the aggressive plan (colored in magenta on the bottom of each subfigure) is less favorable than the normal plan (colored in magenta on the top of each sub-figure), as it leads to a near collision with the object agent, as indicated by the diamond markers. \modelnamenospace$_P$ generates predictions (in olive), which indicate a near collision for the aggressive plan and no collisions for the normal plan, to help the planner choose the correct decision. TAP generates predictions that indicate higher utility for the aggressive plan, leading to the wrong decision.}
    \label{fig:planning_examples}
\vspace{-6mm}
\end{figure}

In Fig.~\ref{fig:planning_examples}, we present a representative scenario to demonstrate the advantage of our model compared to the task-agnostic baseline. In this scenario, the planner proposes two candidate plans\footnote{We omit the conservative plan in the discussion for the sake of simplicity.} (in magenta) for the ego agent, whose observed trajectory is in red: one \textit{normal} plan that yields to the object agent, and one \textit{aggressive} plan that speeds up. For each plan, we visualize the simulated reactive future trajectories of the object agent in cyan and its observed trajectory in blue. In this example, the normal plan is favored over the aggressive plan, as the latter leads to a near collision. 
The predictions (in olive) of our predictor and the baseline are visualized in Fig.~\ref{fig:planning_examples}(a) and (b), respectively. In Fig.~\ref{fig:planning_examples}(b), the task-agnostic predictor TAP generates predictions that indicate near collisions for both plans, which lead to a higher utility for the aggressive plan as it travels further. In contrast, in Fig.~\ref{fig:planning_examples}(a), our predictor \modelnamenospace$_P$ generates predictions that help better approximate the utility for each plan -- the normal plan comes with a higher utility as no near collision is detected, and the aggressive plan results in a lower utility due to a near collision indicated by the predictions. As a result, \modelnamenospace$_P$ helps find the correct decision to choose the normal plan.

\subsection{Example Task 2: Warning for Parallel Autonomy}

\subsubsection{Model Details}
The task-informed predictor model for the warning task leverages the same structure as described in Sec.~\ref{sec:plan_model}, except that it does not include a task information encoder (e.g. the predictions are conditioned only on past observations). In addition, the utility defined in Eq.~\eqref{eq:warn} is not differentiable due to the Boolean comparison operation. So we utilize a soft warning score using the sigmoid function:
\begin{equation}
    r^{(k)} = \text{sigmoid} \big(d_{\text{warn}} - \min_{t=1\ldots T_f} ||x_{\text{ego},t}^{(k)} - x_{\text{object},t}^{(k)}||_2\big).
\label{eq:warn_soft}
\end{equation}
The soft score is close to 1 when the closest distance is smaller than the safety distance threshold, and close to 0 otherwise. We use the same distance threshold of 3.64 meters as in the planning task.

\subsubsection{Quantitative Results}
We evaluate the performance of our model using the same accuracy and task metrics as in the planning task. In the following, we present a series of experiments to validate the advantage of our model.

\textbf{Trade off accuracy for better task performance}
We perform a study on the trade-off between prediction accuracy and task performance, by tuning the coefficient of the task loss $\alpha$ in Eq.~\eqref{eq:loss}.
The results are reported in Table~\ref{tab:warning}.
When $\alpha$ is 0, the model is equivalent to the Waymo baseline~\cite{ettinger2021waymo} as a task-agnostic predictor (TAP) that is optimized only for accuracy. This model achieves the best accuracy metrics. 
As $\alpha$ increases, the task performance improves at the cost of prediction accuracy.
We use $\alpha=20$ in the rest of the experiments, as it achieves a good balance between accuracy and task performance. In practice, the choice of $\alpha$ depends on the specific task requirement.

We further compare the performance of our model to an interactive prediction model M2I~\cite{sun2022m2i} that achieves the state-of-the-art performance in the Waymo benchmark, which yields a minADE of 3.79 meters, a minFDE of 8.40 meters, and an AUC-ROC score of 0.362. The comparison shows that our model is able to achieve much better task performance, i.e. at $\alpha =20$, despite using a simple prediction backbone.

\begin{table}[t!]
\vspace{2mm}
\centering
\begin{tabular}{c|ccccc}
\hline
$\alpha$ & minADE$\downarrow$ & minFDE$\downarrow$ & wADE$\downarrow$ & wFDE$\downarrow$ & AUC-ROC$\uparrow$  \\ \hline
0     & \textbf{4.00} &	\textbf{10.57} & \textbf{7.06} &	\textbf{20.24}    &		0.165 \\
1     & 4.05 &	10.68 & 7.13 &	20.46 &	0.299 \\
5     & 4.19 & 10.98 & 7.32 & 20.92 & 0.449 \\
20     & 4.65 &	11.43 & 7.87 &	21.25 &	0.655 \\
100     & 5.29 & 12.01	& 12.42 & 26.46 &	\textbf{0.776} \\ \hline
\end{tabular}
\caption{Performance of \modelnamenospace$_W$ as a function of the task loss coefficient $\alpha$. Task performance significantly improves at the cost of prediction accuracy.}
\label{tab:warning}
\vspace{-4mm}
\end{table}


\begin{figure}[t!]
    \centering
    \includegraphics[width=0.5\textwidth]{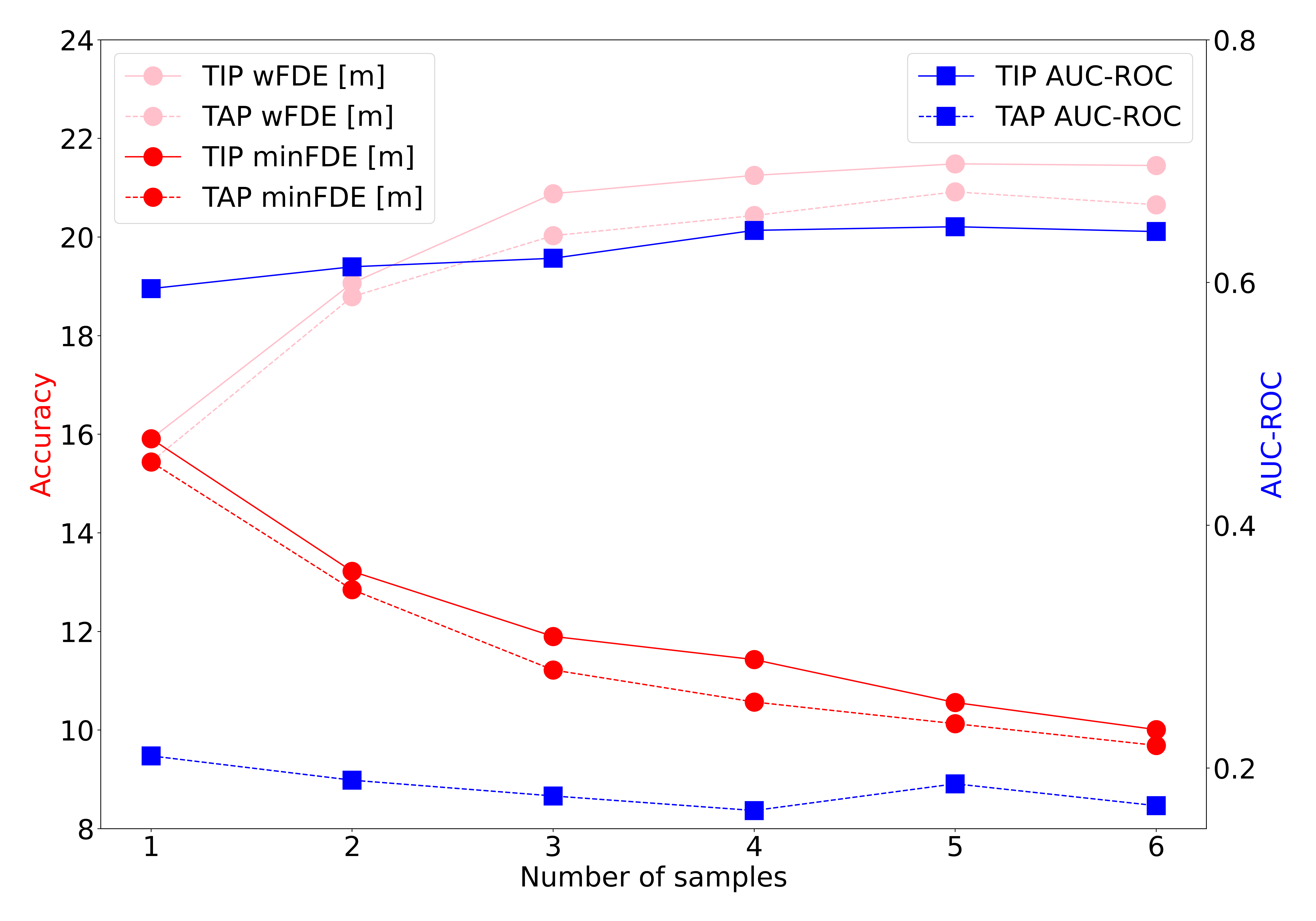}
    \caption{Prediction accuracy (minFDE and wFDE) and warning task performance (AUC-ROC) as a function of the number of prediction samples $K$ from \modelnamenospace$_W$ (solid lines) and the baseline TAP (dashed lines). Compared to TAP, \modelnamenospace$_W$ achieves much better task performance across all samples at the cost of little accuracy. The task performance for \modelnamenospace$_W$ converges at $K = 4$, indicating its predictions serve as approximate sufficient statistics for the warning task with only a few samples.}
    \label{fig:sample_comparison}
\vspace{-5mm}
\end{figure}

\textbf{Support task with limited samples} We examine the performance of our model by varying the number of prediction samples. As depicted in Fig.~\ref{fig:sample_comparison}, our proposed model \modelnamenospace$_W$ achieves much better task performance, by sacrificing little accuracy at different numbers of samples.
For instance, at 4 samples, it improves the AUC-ROC score by more than 3 times at the cost of 8.13\% minFDE (and 5.00\% wFDE) loss compared to TAP, which produces predictions without considering the effects on downstream tasks. Besides, the task performance for \modelnamenospace$_W$ stabilizes at 4 samples. This demonstrates that \modelnamenospace$_W$ provides predictions that summarize the influence by nearby agents as approximate sufficient statistics for the warning task using only a few samples. 


\textbf{Robust to noisy utility estimates}
We further investigate how robust our predictor is when trained with noisy utility estimates, to probe our approach under imperfect information about the task. At training time, we add to the estimated utilities a random Gaussian noise with zero mean and increasing variance levels and observe that the task performance decreases slightly by 3.08\%, within a noise level of 25\% of the magnitude of the utility, compared to a standard \modelnamenospace$_W$ model that is trained without noise. The observation demonstrates that our approach is robust to perturbations in the utility estimates during training, as an example of imperfect task information, or an example of adapting to a task with a slightly different utility function.

\textbf{Handle multiple object agents}
We performed additional experiments by training \modelnamenospace$_W$ to predict joint future trajectories of up to 4 agents, including 3 object agents. Compared to the TAP baseline that has a minFDE of 10.03 meters and an AUC-ROC score of 0.226, \modelnamenospace$_W$ achieves a much higher task score of 0.314 with a slightly worse minFDE of 10.57. This demonstrates that our predictor works well for more than two agents. Full results are available in Table~\ref{tab:warning_8}.

\subsubsection{Qualitative Results}
\begin{figure}[t!]
\vspace{2mm}
    \centering
    \includegraphics[width=0.4\textwidth]{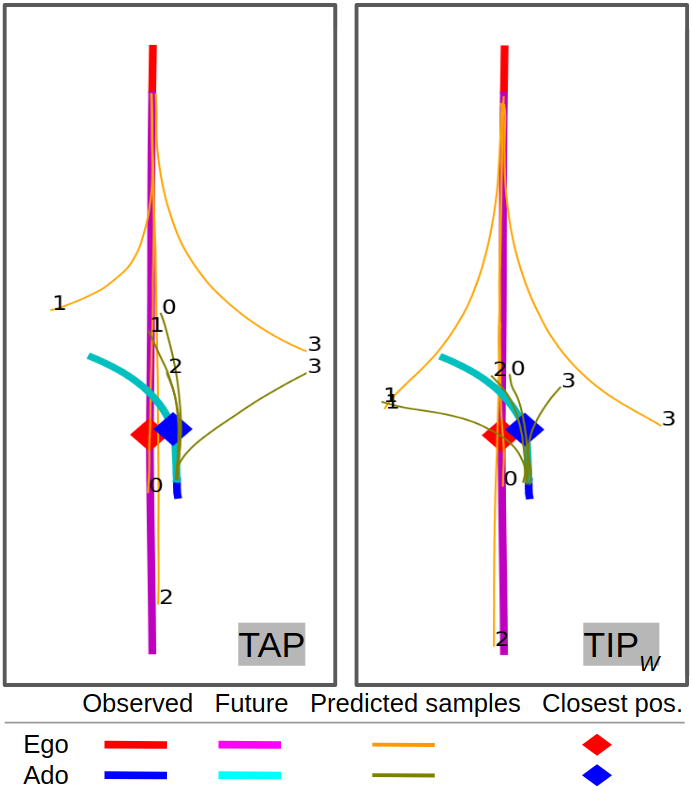}
    \caption{Comparison between TAP (left) and \modelnamenospace$_W$ (right) in a warning scenario, where the closest distance is indicated by the diamond markers. The sample indices are annotated to help associate joint predictions. Our model \modelnamenospace$_W$ generates predictions (in olive and orange) that help identify multiple instances of near collisions, especially through joint samples \#2 that match with the observed future trajectories.}
    \label{fig:warning_example}
\vspace{-4mm}
\end{figure}

We present a representative warning example in Fig.~\ref{fig:warning_example}, where the two agents are getting too close according to their observed future trajectories. The closest distance is indicated by the diamond markers. The sample index $k$ is labelled to help identify joint predictions. From the left plot, we see that the task-agnostic baseline fails to identify a likely near collision. Although it predicts a near collision with joint samples \#3, they are predicted with a very low probability, i.e. smaller than 1\%. In contrast, our predictor on the right identifies multiple near collision instances, especially through samples \#2 that successfully predicts the object agent is going to cross the ego agent's path, which matches with what happens in the data. As a result, its predictions indicate a high likelihood of collision and lead to the correct warning decision. This example also verifies that in the downstream task, it is usually not required to generate \textit{perfect} predictions, as long as the predictions cover sufficient statistics for the task (e.g. the predictions indicate a collision).

\vspace{-2mm}
\section{CONCLUSION}
\vspace{-1mm}
We propose a task-informed motion prediction system, in which predictors are trained to both make accurate predictions and support correct decision making in a downstream task. By leveraging a specification of the task, we allow the predicted samples to provide approximate sufficient statistics of the environment for the task and support a variety of tasks, without requiring a full differentiable task for co-training. We demonstrate our predictor in two tasks on the Waymo dataset, and show its advantage through quantitative and qualitative experiments. Future work includes further improving the task performance through a stronger backbone model and performing experiments in additional benchmarks.

\bibliographystyle{IEEEtran}
\bibliography{references} 

\newpage
\section*{APPENDIX}
\subsection{Additional Experiments in Planning Task}
In this section, we present additional experiments in the planning task that demonstrate our model works with an additional simulation system.

\subsubsection{IDM Simulation Model}
\begin{table}[t!]
\vspace{2mm}
\centering
\begin{tabular}{c|cc}
\hline
Model & AUC-ROC$_P$$\uparrow$ & AUC-ROC$_{Pa}$$\uparrow$  \\ \hline
TAP  & 0.732 & 0.680 \\
TIP$_P$   &	\textbf{0.802} & 0.620 \\
TIP$_{Pa}$    &	0.623 & \textbf{0.825} \\ \hline
\end{tabular}
\caption{Comparison between our TIP$_P$ and the baseline model TAP, in terms of the task performance, by using an \textbf{IDM} model as in BARK to simulate realistic ado agent behaviors. We observe consistent improvements in task performance with a different simulation model.}
\label{tab:planning_idm}
\end{table}

We performed additional experiments by adopting the IDM model used in BARK ~\cite{bernhard2020bark} to simulate ado agent behaviors. The results are summarized in Table~\ref{tab:planning_idm}. We observe that our proposed approach shows improved performance in the task performance under a different simulation model, which is consistent with what we observe in Table~\ref{tab:planning}. 

\subsection{Additional Experiments in Warning Task}
In this section, we present additional experiments in the warning task that demonstrate our model works for more than two agents.

\subsubsection{Multiple Object Agents}
We performed additional experiments in the warning task by using our proposed predictor to predict \emph{joint} future trajectories for 4 agents, including the ego agent and 3 closest object agents in each scenario. The warning signal is triggered if any of the object agents is getting too close to the ego vehicle trajectory, using the same criteria as in Eq.~\eqref{eq:warn}. The comparison is summarized in Table~\ref{tab:warning_8}. 
The results show that our proposed predictor is able to improve the task performance by a large margin (39\%), at the cost of little accuracy (5.4\% for minFDE), with a reasonable $\alpha$ value of 20, compared to the task-agnostic baseline ($\alpha$ = 0). 

\begin{table}[t!]
\vspace{2mm}
\centering
\begin{tabular}{c|ccccc}
\hline
$\alpha$ & minADE$\downarrow$ & minFDE$\downarrow$ & wADE$\downarrow$ & wFDE$\downarrow$ & AUC-ROC$\uparrow$  \\ \hline
0     & \textbf{6.16} &	\textbf{10.03} & \textbf{7.34} & \textbf{13.48}    &		0.226 \\
1     & 6.21 &	10.18 & 7.43 & 13.60 &	0.232 \\
5     & 6.44 & 10.46 & 7.57 & 13.81 & 0.251 \\
20     & 6.59 &	10.57 & 8.52 & 15.07 &	0.314 \\
100     & 7.32 & 11.52 & 14.51 & 22.48 & \textbf{0.477} \\ \hline
\end{tabular}
\caption{Performance of TIP$_W$ as a function of the task loss coefficient $\alpha$ when \textbf{predicting for 4 agents}. Task performance significantly improves at the cost of prediction accuracy.}
\label{tab:warning_8}
\end{table}

\end{document}